# Detecting weak and strong Islamophobic hate speech on social media

Bertie Vidgen[1]
Oxford Internet Institute, University of Oxford
bertievidgen@gmail.com

Taha Yasseri
Oxford Internet Institute, University of Oxford
Alan Turing Institute, London
taha.yasseri@oii.ox.ac.uk

## ABSTRACT

Islamophobic hate speech on social media inflicts considerable harm on both targeted individuals and wider society, and also risks reputational damage for the host platforms. Accordingly, there is a pressing need for robust tools to detect and classify Islamophobic hate speech at scale. Previous research has largely approached the detection of Islamophobic hate speech on social media as a binary task. However, the varied nature of Islamophobia means that this is often inappropriate for both theoretically-informed social science and effectively monitoring social media. Drawing on in-depth conceptual work we build a multi-class classifier which distinguishes between non-Islamophobic, weak Islamophobic and strong Islamophobic content. Accuracy is 77.6% and balanced accuracy is 83%. We apply the classifier to a dataset of 109,488 tweets produced by far right Twitter accounts during 2017. Whilst most tweets are not Islamophobic, weak Islamophobia is considerably more prevalent (36,963 tweets) than strong (14,895 tweets).

Our main input feature is a gloVe word embeddings model trained on a newly collected corpus of 140 million tweets. It outperforms a generic word embeddings model by 5.9 percentage points, demonstrating the importan4ce of context. Unexpectedly, we also find that a one-against-one multi class SVM outperforms a deep learning algorithm.

## 1 Introduction

In recent times, the prevalence, effects and spread of Islamophobic hate speech on social media has received considerable attention from government (APPG on British Muslims, 2018; HM Government, 2012; Home Affairs Select Committee, 2017), Muslim community groups (Ingham-Barrow, 2018; Runnymede Trust, 2017; Tell Mama, 2018), academics (Allen, 2010; Burnap & Williams, 2016) and the platforms themselves (Facebook, 2018; Twitter, 2018). Islamophobic hate speech inflicts considerable harm on both targeted individuals and wider society, and risks reputational damage for the host platforms.

Islamophobia has been variously described as a form of racism (Meer & Modood, 2009), stereotyping (Moosavi, 2015), prejudice (Imhoff & Recker, 2012), fear (Kunst, Sam, & Ulleberg, 2013), exploitation (Beck, Charania, & Al-issa, 2017), exclusion (Bayrakli & Hafez, 2018) and dominance (Jackson, 2018). It can be understood as what W. B. Gallie terms an 'essentially contested concept' – a concept with numerous definitions and descriptions but little consensus as to what the core features are (Gallie, 1956). In the present work, we use Bleich's widely cited definition of Islamophobia, as this reflects much other work undertaken by leading theorists in the field (Allen, 2010; Awan, 2016; Ekman, 2015). Bleich defines Islamophobia as: 'Indiscriminate negative attitudes or emotions directed at Islam or Muslims.' (Bleich, 2011, p. 1581). This definition can be adapted for social media posts:

> "Any content which is produced or shared which expresses indiscriminate negativity against Islam or Muslims."

Recent interventions in social psychology point to the multifaceted nature of prejudice; from behaviors which are explicit, overt and direct to those which are implicit, covert and indirect (Nadal, Griffin, Hamit, Leon, & Rivera, 2012; Pettigrew & Meertens, 1995). Thus, in recent times much research has focused on 'everyday' prejudicial and hateful actions (Dunn & Hopkins, 2016; Moosavi, 2015) as well as 'micro-aggressions' (Haque, Tubbs, Kahumoku-Fessler, & Brown, 2018; Husain & Howard, 2017). Little research has explicitly explored these distinctions with regard to hate speech. This is surprising given that distinguishing between different types of Islamophobic speech offers considerable empirical and theoretical advantages over using a single category of 'Islamophobia'. It enables researchers to better understand the dynamics of Islamophobia (which may differ across different manifestations) and to investigate radicalization processes, whereby individuals progress from being weak to strongly Islamophobic. It is also important for enabling platforms and governments to better regulate and monitor social media and provide support to victims. Ultimately, it could also lead to better detection of hate speech; Waseem and Hovy note that 'in much hate speech research, diverse types of abuse have been lumped together under a single label, forcing models to account for a large amount of within-class variation.' (Waseem & Hovy, 2016, p. 82).

---

[1] Primary author, email for further information.

## 2 Classification task

The classification task addressed is to distinguish between non-Islamophobic, weak Islamophobic and strong Islamophobic social media content. Drawing on Bleich's definition of Islamophobia, as well as work undertaken with victims of Islamophobia by the Runnymede Trust, MIND and Tell Mama (Ingham-Barrow, 2018; Runnymede Trust, 2017; Tell Mama, 2018), we define strong Islamophobic hate speech as:

> Speech which explicitly expresses negativity against all Muslims.

This can vary, from expressing explicitly negative *views*, such as describing Muslims as barbarians to calling for prejudicial *actions*, such as demanding that Muslims are forcibly banned from the UK. We define weak Islamophobic hate speech as:

> Speech which weakly expresses negativity against all Muslims
>
> AND
>
> Speech which explicitly expresses negativity against a specific subset of Muslims

An example of the first type of Islamophobia is making a comment about how Muslims are 'different' or have unusual cultural practices. An example of the second type of weak Islamophobia is sharing a new story about a terrorist attack and explicitly foregrounding the fact that the perpetrator is a Muslim.

Whereas blatant Islamophobia is easy to spot, subtle Islamophobia is often harder to observe and may only partially manifest anti-Muslim negativity.

## 3 Previous work

Previous work in this area demonstrates the challenges of – but also potential for – creating a classification system which distinguishes between weak and strong Islamophobic hate speech. To the authors' best knowledge, no previous research has focused specifically on this task (Schmidt & Wiegand, 2017). Nonetheless, many prior studies are still relevant to the present discussion, not least because most of them have used data from the same source (Twitter). Most previous research has also focused on binary rather than multi-class classification. Classifier performance in the latter task is often far lower. As Salminen et al. note in a recent paper, 'existing works using multi-label classification for online hate speech are extremely rare, and we could not locate prior work that had achieved good results.' (Salminen et al., 2018, p. 331)

Most existing research into multi-class classification has focused on distinguishing between different *targets* of hate rather than different *strengths* (Burnap & Williams, 2016; Park & Fung, 2017; Saleem, Dillon, Benesch, & Ruths, 2017; Salminen et al., 2018; Silva, Mondal, Correa, Benevenuto, & Weber, 2016). It is difficult to complete both tasks at once; distinguishing between different strengths of hate inevitably involves narrowing the domain to just one target (here, Islamophobia )as 'hateful speech classification systems require target-relevant training' (Saleem et al., 2017, p. 7) Classifying content based on strength rather than target poses additional challenges as there is less variation between classes; weak and strong Islamophobic tweets often use similar keywords, grammatical structures and non-verbal features (such as embedded hyperlinks and emojis).

Burnap and Williams train a classifier to distinguish between different levels of cyberhate (divided into 'moderate' and 'extreme' classes) targeted against Black Minority Ethnic (BME) and religious groups on Twitter (Williams & Burnap, 2016), achieving precision of 0.77.

Malmasi and Zampieri distinguish between 'hate' speech, 'Offensive' speech and 'Ok' speech. They achieve 78% accuracy but on an unevenly weighted training/testing dataset – over half of their corpus is 'OK'. Their model struggles to distinguish between non-OK content; of 2,399 'Hateful' instances in their dataset, 1,050 are categorised correctly, 1,113 are miscategorised as 'Offensive' and 236 as 'OK'. They also do not test their model on unseen data, only reporting the results of cross-validation (Malmasi & Zampieri, 2017).

Jha and Mahmidi distinguish between 'benevolent' and 'hostile' sexism. They use Waseem and Hovy's dataset of 16,000 tweets as well as ~7,000 newly collected ones (Waseem & Hovy, 2016). Using SVM they report an F1 score of 0.80 for Benevolent tweets, 0.48 for Hostile and 0.89 for Others. A their data is highly skewed towards Others rather than Hostile, overall performance is strong.

Kumar et al (Kumar, Ojha, Malmasi, & Zampieri, 2018) distinguish between overtly aggressive, covertly aggressive and non-aggressive tweets using a dataset of 15,000 Facebook posts. In a competition entered by 130 teams (of which 20 completed it and provided the technical details of their model), the highest performing obtained a weighted F-score of 0.64. As the authors note, 'the results […] depict how challenging the task is.' (Kumar et al., 2018, p. 1)

Davidson et al. train a model to distinguish between hate speech and offensive speech, and non-offensive speech in tweets. They report impressive results, with precision of 0.91, recall of 0.90 and an F1 score of 0.90. Their work demonstrates the potential for multiclass classification, makes an important theoretical argument apropos the need to separate different types of content, and introduces the use of 'Ease of Reading' metrics as an input feature. However, as they note, their model performs poorly with hate speech, of which almost 40% is misclassified. The high F1 score is largely due to the fact that their classes are very uneven (76% of the data is in the 'offensive speech' category). They also train and test their classifier on a single dataset, which could risk overfitting.

## 4 Data

We collect a new dataset of 140 million tweets produced by Twitter followers of mainstream and far right UK political parties. The data is collected over the course of 2017 and the first six months of 2018.

It also includes some tweets from before 2017 which are made available by Twitter's Rest API. The tweets come from:

1. 7,500 users randomly selected from followers of UKIP.
2. 7,500 users randomly selected from followers of the Conservatives.
3. 7,500 users randomly selected from followers of Labour.
4. 7,500 users randomly selected from followers of the Liberal Democrats.
5. All ~15,000 followers of the BNP.
6. All ~32,000 followers of Britain First.
7. Every tweet produced by a set of 45 far right accounts (consisting of every group which appear in Hope Not Hate's 2015 and 2017 reports on the far right, and which have a Twitter account (Hope Not Hate, 2015, 2017)).
8. Every @ mention of the same 45 far right accounts (collected from the Twitter Stream API).

## 4.1 Data annotation

We create a training dataset of 4,000 tweets by sampling from across all 8 of the sources outlined above. Creating a training dataset with sufficient instances of hateful content is a time-consuming endeavor, not least because in most online contexts the prevalence of hate is relatively low overall (Schmidt & Wiegand, 2017, p. 7).To ameliorate this problem, Waseem and Hovy recommend increasing the prevalence of hate speech by sampling data which contains relevant topics (Waseem & Hovy, 2016). This approach is partially adopted here; we sample 1,000 tweets using keyword searches for 'Muslims' and 'Islam'.

All 4,000 tweets are annotated blind by three annotators who are experts in UK politics and the study of prejudice. The annotators all use the same annotation guidelines. The guidelines are based on the definition of Islamophobia offered above and were iteratively developed through two preliminary studies, each consisting of 200 tweets. Across the 4,000 tweets, inter-rater agreement is high. Percentage agreement is 89.9%, Fleiss' kappa is 0.837 and Krippendorf's alpha is 0.895. We also compute category-wise scores for Fleiss' kappa, which range from 0.737 for Weak Islamophobia to 0.907 for Strong Islamophobia. The consistency of these results show the robustness of the annotation guidelines and how they were implemented.

In cases where annotators disagree, tweets are assigned to classes based on the majority decision. In the final dataset, 3,106 tweets are classed as 'Not Islamophobic', 484 tweets are classed as 'Weak Islamophobic', 410 tweets are classed as 'Strong Islamophobia'. To create an evenly-weighted dataset the number of 'Not Islamophobic' tweets is reduced through random sampling to 447 instances (the difference between the number of tweets in the other two classes). This creates a final dataset of 1,341 tweets.

## 5 Input features

Feature selection refers to the choice of input variables used to train the classifier. In many cases features are selected using 'brute force' computation via a grid search with little consideration for *why* they have been included. Models in which variables are selected without any theoretical justification may perform well in initial testing but risk overfitting, and as such are unlikely to be generalizable, making them unsuitable for empirical research (Domingos, 2012). Thus, it is crucial that the model is not only accurate but that its choice of inputs can be explained and thus avoids becoming a 'black box' (Biran & McKeown, 2017). Accordingly, in the present work, we only consider features which can be theoretically justified.

First, we create a text only model, using one-hot encodings for each term. Second, we create a model using 50 surface-level and derived non-text features. These include sentiment and polarity (Feuerriegel & Proellochs, 2018), count of swear words (Ipsos MORI, 2016) and parts of speech and named entities (Benoit & Matsuo, 2018). We also derive two new input features, mentions of Muslim names and mentions of Mosques, both taken from relevant Wikipedia pages. Third, we create a combined model that uses both one-hot encodings and all 50 of the non-text features. Fourth, we create a model using pre-trained gloVe word embeddings, trained on two billion tweets (Stanford, 2018). Fifth, we create a gloVe model using newly-trained word embeddings on the corpus of 140 million tweets (Pennington, Socher, & Manning, 2014). Finally, sixth, we create a model which uses the newly-trained word embeddings as well as all 50 of the non-text features.

For testing we implement ten-fold cross-validation on the Naïve Bayes algorithm as previous research indicates that it generally outperforms most other off-the-shelf algorithms for text classification tasks (Kotsiantis, 2007; Wainer, 2016; Wang & Manning, 2012) and it is deterministic, producing the same results each time it is implemented. The results are shown in Table 1.

| Input feature model | Accuracy |
|---|---|
| Model 1: Text only (one-hot encoding) | 30.07% |
| Model 2: Non-text features | 49.96% |
| Model 3: Text + non-text features | 30.36% |
| Model 4: Pre-trained word embeddings | 63.20% |
| Model 5: Newly trained word embeddings | 69.13% |
| Model 6: Newly trained word embeddings + all non-text features | 65.20% |

**Table 1: Accuracy of models with different input features**

The best performing model is the newly trained word embeddings alone (model 5). Interestingly, this considerably outperform the accuracy of the pre-trained word embeddings model (5.9 percentage points, 69.13% compared with 63.2%). This suggests that the benefits of having tweets which are contextually-specific outweighs the cost of having a smaller dataset. This is in line with previous work, such as Lai et al., who report that 'corpus domain is more important than corpus size.' (Lai, Liu, He, & Zhao, 2016, p. 8). We optimize the newly trained word embeddings model by including additional non-text features, testing for up to ten additional features through an exhaustive grid search. We find that the count of mentions of Mosques is consistently an important input feature, suggesting that this newly engineered feature could also be used in other studies. The final model (model 7), which maximizes accuracy, contains 6 additional non-text features:

> Word embeddings + count of mentions of Mosques + presence of HTML + presence of RT + part of speech: 'conjunction' + named entity recognition: 'location' + named entity recognition: 'organization'

## 6   Choice of algorithm

We test the newly trained word embeddings model (model 5 in Table 1) on six different algorithms on, selected based on previous research on classification (Kotsiantis, 2007; Wainer, 2016; Wang & Manning, 2012): Naïve-Bayes, Random Forests (with trees = 10, 100 and 1,000), Logistic Regression, Decision Trees, SVM and Deep Learning. We implement multi-class SVM with a one-against-one strategy (Hsu & Lin, 2002). Through an exhaustive grid search we optimize the hyperparameters of the SVM classifier with a 'radial' kernel. 'C' is 2 and gamma is 0.01. We also optimize the Deep Learning model, testing for the activation function, optimization function, learning rate and number of epochs. The results, including optimized hyperparameters, are shown in Table 2.

| Algorithm | Accuracy |
|---|---|
| Naïve-Bayes | 69.13% |
| Random Forests (trees = 10) | 65.40% |
| Random Forests (trees = 100) | 68.72% |
| Random Forests (trees = 1000) | 67.94% |
| Logistic Regression | 69.13% |
| Decision Trees | 61.23% |
| SVM with kernel = 'radial' + 'C' = 2 + gamma = 0.01 | 72.17% |
| Deep Learning with epochs = 100 + activation function = 'relu' + optimization function = rmsprop, learning rate = 0.001 | 71.14% |

**Table 2: Results of algorithm testing**

All six algorithms perform well, with accuracy ranging from 61.23% to 72.17%. The two highest performing are SVM and Deep Learning (using only a feed forward 'shallow' architecture) – the accuracy of SVM is 72.17%, which outperforms Deep Learning by 1.03 percentage points. Thus, contrary to our initial expectations, we opt to use SVM for the classifier. The performance of SVM and Deep learning algorithms for text classification has long been a point of debate within machine learning (Zaghloul, Lee, & Trimi, 2009). Although Deep Learning has been heralded as the future of machine learning, several recent studies suggest that SVM can outperform it in certain applications (Korba & Arbaoui, 2018; Liu, Choo, Wang, & Huang, 2017). Our result contributes to ongoing discussions in this area.

The SVM hyperparameters are set to maximize generalizability (i.e. low 'C' and gamma values), which make the classifier suitable for empirical applications.

## 7   Performance

### 7.1   Cross-validated performance

The classifier consists of model 7 implemented with a tuned SVM. We cross-validate the classifier on the training data set (n = 1,341 tweets) using ten-fold classification. The results are shown in Table 3.

| Fold | Accuracy | Balanced accuracy | Precision | Recall | F1 score |
|---|---|---|---|---|---|
| 1 | 0.796 | 0.846 | 0.795 | 0.798 | 0.797 |
| 2 | 0.76 | 0.808 | 0.75 | 0.736 | 0.743 |
| 3 | 0.736 | 0.808 | 0.74 | 0.75 | 0.745 |
| 4 | 0.721 | 0.792 | 0.714 | 0.724 | 0.719 |
| 5 | 0.718 | 0.774 | 0.686 | 0.685 | 0.686 |
| 6 | 0.746 | 0.808 | 0.74 | 0.742 | 0.741 |
| 7 | 0.702 | 0.785 | 0.699 | 0.721 | 0.71 |
| 8 | 0.79 | 0.845 | 0.793 | 0.793 | 0.793 |
| 9 | 0.756 | 0.809 | 0.736 | 0.736 | 0.736 |
| 10 | 0.735 | 0.798 | 0.733 | 0.729 | 0.731 |
| Mean | 0.746 | 0.807 | 0.739 | 0.741 | 0.740 |

**Table 3: Performance of classifier over ten folds**

For the accuracy, recall and precision scores (and, as such, F1 scores) we use the macro-aggregation strategy described by Sokolova and Lapalme, in which values are calculated for each class and then the per-class agreement is averaged, with each class treated equally (Sokolova & Lapalme, 2009). The classifier performs similarly for recall and precision (0.741 and 0.739 respectively), and as such has a comparable F1 score (0.74). This is encouraging as it means that the classifier does well at balancing the need to identify relevant instances with minimizing misclassifications, and as such can be applied to real world 'wild' data. We also test for balanced accuracy. This is a relatively new metric put forward by Velez et al. which combines specificity and sensitivity (Velez et al., 2007). They argue that it helps to overcome imbalanced classes and is well-suited to smaller datasets where even small differences in class size can have considerable impact. We report high balanced accuracy (0.807), which provides further evidence that the classifier does well at balancing identifying relevant instances with minimizing misclassifications.

## 7.2 Performance on unseen data

To check the classifier's performance in 'the wild' we apply it to an unseen dataset of 109,488 tweets produced by 45 far right Twitter accounts during 2017. 100 tweets are randomly sampled from tweets assigned to each of the three classes (None, Weak Islamophobia and Strong Islamophobia) to create a new combined dataset of 300 tweets. This is annotated blind by the three annotators who annotated the original training dataset, using the same annotation guidelines. As before, we take the majority decision to decide the annotation (in 95% of cases all three annotators are in perfect agreement). The results of this testing, as well as how it compares with the previous 10-fold testing, are shown in Table 4. Interestingly, the classifier performs better across all metrics on the unseen data, with accuracy of 77.3%. The uplift in performance, and consistency of the results, indicates the robustness of our approach and its generalizability, which is most likely due to our selection of theoretically-informed input features. Importantly, these results suggest that the classifier is suitable for implementation in empirical research as performance is well above the 70% minimum precision recommended by van Rijsbergen (van Rijsbergen, 1979).

|  | Accuracy | Balanced accuracy | Precision | Recall | F1 score |
|---|---|---|---|---|---|
| **Results on unseen data** | 0.773 | 0.83 | 0.778 | 0.773 | 0.776 |
| **Difference with ten-fold testing** | 0.027 | 0.023 | 0.039 | 0.032 | 0.036 |

**Table 4: Performance of classifier on unseen data**

The classifier performs well at distinguishing None Islamophobic from Strong Islamophobic. However, it struggles with distinguishing Weak from both Strong and None. For instance, out of 100 tweets which are labelled as Strong Islamophobic, 23 are actually Weak. Similarly, out of 100 predicted Weak Islamophobic tweets, 22 are actually None. This is shown in Figure 1, a contingency table of the classifier's performance on unseen data.

Qualitative investigation of the 300 tweet dataset shows that, in many cases, the None Islamophobic tweets express hatred and prejudice against other groups, such as immigrants. Some also discuss Muslims and Islamic practices but without expressing any negativity. Distinguishing between instances such as these is a challenge as they often have similar input features.

|  |  | Predicted Islamophobia | | | |
|---|---|---|---|---|---|
|  |  | None | Weak | Strong |  |
| **Actual** | None | 91 | 22 | 4 | 117 |
|  | Weak | 8 | 68 | 23 | 99 |
|  | Strong | 1 | 10 | 73 | 84 |
|  |  | 100 | 100 | 100 | **300** |

**Figure 1: Contingency table for performance on unseen data**

## 8 Application to far right tweets

To demonstrate the utility of distinguishing between different classes of Islamophobic hate, we show the results of applying the classifier to the 109,488 tweets produced by 45 far right accounts. This is in Figure 2. Noticeably, whilst most tweets are not Islamophobic (57,630 tweets), weak Islamophobia is considerably more prevalent (36,963 tweets) than strong Islamophobia (14,895 tweets). In future empirical research, the classifier could be used to better understand the dynamics of these respective types of Islamophobic hate speech, such as how they fluctuate over time.

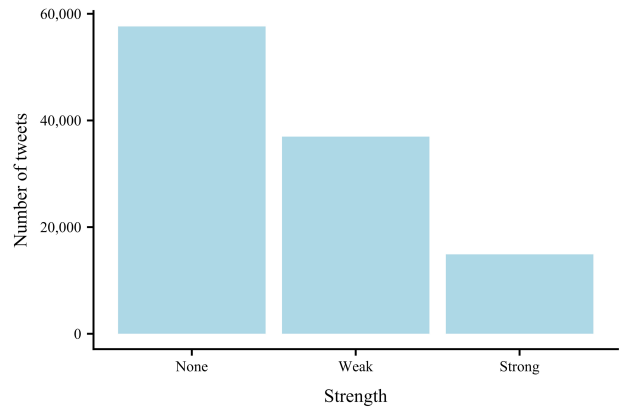

**Figure 2: Prevalence of tweets for different strengths of Islamophobia**

## 9 Conclusion

The multi-class Islamophobic hate speech classifier developed in the present work marks an important step forward in developing quantitative methods to provide detailed insight into online Islamophobia. The findings are also relevant for classifying and studying other forms of hate, such as misogyny, racism and anti-Semitism. Whilst more work needs to be undertaken, particularly in making nuanced distinctions between different strengths of hate, the results reported here are promising (particularly, accuracy of 77.3% and balanced accuracy of 83%). In our future work we plan on improving the classifier's performance by increasing the size of the training dataset and engineering additional input features.